\pdfoutput = 1

\documentclass{article}

\usepackage{authblk}
\usepackage{algorithm}
\usepackage{algpseudocode}
\usepackage{amsmath}
\usepackage{amsthm}
\usepackage{graphicx}
\usepackage{appendix}
\usepackage{setspace}

\begin{document}
\title{Robust Subspace Recovery via Bi-Sparsity Pursuit}
\author{Xiao Bian, Hamid Krim}
\affil{North Carolina State University}
\date{}
\maketitle

\begin{abstract}
Successful applications of sparse models in computer vision and machine learning~\cite{elad2010role}\cite{elad2012sparse}\cite{elham2009sparse} imply that in many real-world applications, high dimensional data is distributed in a union of low dimensional subspaces. Nevertheless, the underlying structure may be affected by sparse errors and/or outliers. In this paper, we propose a bi-sparsity model as a framework to analyze this problem and provide a novel algorithm to recover the union of subspaces in presence of sparse corruptions. We further show the effectiveness of our method by experiments on both synthetic data and real-world vision data. 

\end{abstract}

\section{Introduction}

Separating data from errors and noise has always been a critical and important problem in signal processing, computer vision and data mining~\cite{elad2012sparse}. Robust principal component pursuit is particularly successful in recovering low dimensional structures of high dimensional data from arbitrary sparse outliers~\cite{RPCA}. Successful applications of sparse models in computer vision and machine learning~\cite{elad2010role}~\cite{rubinstein2012} have, however, increasingly hinted at a more general model, namely that the underlying structure of high dimensional data looks more like \emph{a union of subspaces} (UoS) rather than \emph{one low dimensional subspace}. Therefore, a natural and useful extension question is about the feasibility of such an approach in high dimensional data modeling where the union of subspaces is further impacted by outliers and errors. This problem is intrinsically difficult, since the underlying subspace structure is also corrupted by unknown errors, which may lead to unreliable measurement of distance among data samples, and make data deviate from the original subspaces. 

Recent studies on subspace clustering~\cite{liu2013}~\cite{elham2009sparse}~\cite{sol2012} show a particular interesting and a promising potential of sparse models. In~\cite{liu2013}, a low-rank representation (LRR) recovers subspace structures from sample-specific corruptions by pursuing the lowest-rank representation of all data jointly. The contaminated samples are sparse among all sampled data. The sum of column-wise norm is applied to identify the sparse columns in data matrices as outliers. In~\cite{elham2009sparse}, data sampled from UoS is clustered using sparse representation. Input data can be recovered from noise and sparse errors under the assumption that the underlying subspaces are still well-represented by other data points. In~\cite{sol2012}, a stronger result is achieved such that data may be recovered even when the underlying subspaces overlap. Outliers that are sparsely distributed among data samples may be identified as well.

In this paper, we consider a more stringent condition that all data samples may be corrupted by sparse errors. Therefore the UoS structure is generally damaged and no data sample is close to its original subspace under a measure of Euclidean metric. More precisely, the main problem can be stated as follows: 

\newtheorem{myprob}{Problem}
\begin{myprob}
\label{prob1}
Given a set of data samples $X = [x_1,x_2,\dots,x_n]$, find a partition of $X$, such that each part $X_I$ can be decomposed into a low dimensional subspace (represented as low rank matrix $L_I$) and a sparse error (represented as a sparse matrix $S_I$), such that
	$$X_I =  L_I + S_I, I = 1,\dots, J$$
\end{myprob}

\begin{figure}[t]
\label{preFig}
\begin{center}
\begin{tabular}{c}
\includegraphics[width=0.6\textwidth]{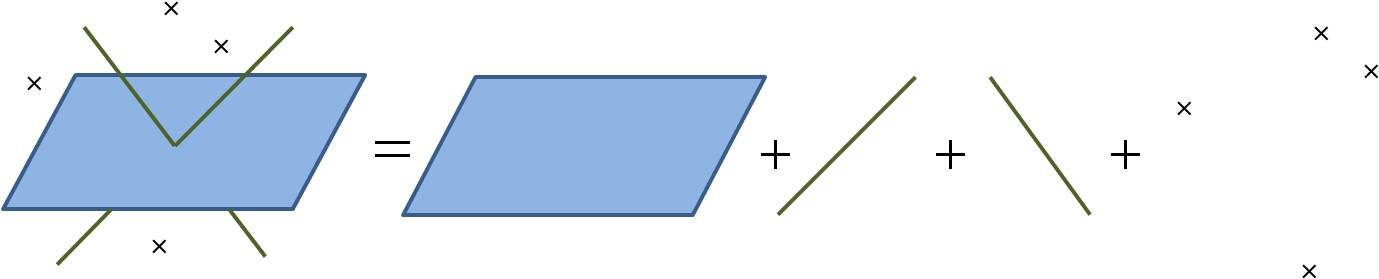} \\
\small (a) Robust Subspace Recovery\\
\includegraphics[width=0.55\textwidth]{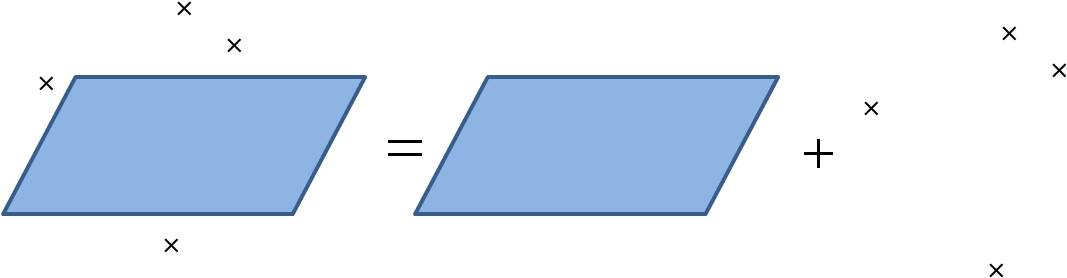} \\
\small (b) Robust Principal Component Analysis\\

\end{tabular}
\end{center}
\caption{Illustration of Problem \ref{prob1}}
\end{figure}

Each $L_I$ then represents one low dimensional subspace of the original data space, and $L = [L_1 | L_2 | \dots | L_J]$ the union of subspaces. Furthermore, the partition would recover the clustering structure of original data samples hidden from the errors $S = [S_1 | S_2 | \dots | S_J]$. 

Concretely, the goal of this problem is twofold: First, we wish to find out the correct partition of data so that data subset reside in a low dimensional subspace. Second, we wish to recover each underlying subspace from the corrupted data. It is worth noting that the corrupted data may highly affect the partition, and hence decoupling the two tasks would be problematic. In this paper, we propose an integral method to decompose the given corrupted data matrix into two parts, representing the clean data and sparse errors, respectively. The correct partition of data, as well as the individual subspaces, are also simultaneously recovered. Moreover, we prove a condition for the data to be exactly recovered as the global minimum of the proposed optimization problem, and provide an algorithm to approximate the global optimizer, which henceforth refer to as Robust Subspace Recovery via Bi-Sparsity Pursuit (RoSuRe). 

\subsection{Organization of the paper}
The remainder of this paper is organized as follows. In Section \ref{ProbForm}, we provide the fundamental concepts necessary for the development of our proper modeling. Building on this model, we reformulate Problem \ref{prob1} as an optimization problem in Section \ref{SRmethod}, and develop the rationale along with the condition for subspace recovery. In Section \ref{algorithm}, we introduce the RoSuRe algorithm for robust subspace recovery. In Section \ref{experiments}, we finally present experimental results on synthetic data and real-world applications. 

\subsection{Notation}
A brief summary of notations used throughout this paper is follows: The dimension of a $m\times n$ matrix $X$ is denoted as $dim(X) = (m,n)$. $\|X\|_0$ denotes the number of nonzero elements in $X$, while $\|X\|_1$ same as the vector $l_1$ norm.  For a matrix $X$ and an index set $J$, we let $X_J$ be the submatrix containing only the columns of indices in $J$. $col(X)$ denotes the column space of matrix $X$. We write $P_{\Omega_A}X$ as the orthogonal projection of matrix $X$ on the support of $A$, and $P_{\Omega_A^c}X = X - P_{\Omega_A}X$. The sparsity of a $m\times n$ matrix $X$ is denoted by $\rho(X) = \frac{\|X\|_0}{mn}$.

\section{Problem Formulation}
\label{ProbForm}
\subsection{A union of subspaces with corrupted data}

Consider a data set $l\in R^d$ uniformly sampled from a union of subspaces $S = \cup_{i=1}^JS_i$, then assumed sufficient sample density, each sample can be represented by the others from the same subspace with probability 1.\footnote{All hyperplanes of a subspace here are of measure 0. Therefore the distribution of samples will span the entire subspace with probability 1. } Mathematically, we represent the data matrix by $L = [l_1|l_2|\dots|l_n]$ , yielding
$$
L = LW,
$$
where $W$ is $n\times n$ block-diagonal matrix. 

More specifically, let $n_i$ be the number of samples from $S_i$, and $b_i$ the dimension of block $W_I$ of $W$, then $n_i \geq b_i$. It follows that $b_i \leq \max_i\{n_i\}$. This condition constrains $W$ to be a sparse matrix, since $\rho(W) = \|W\|_0/n^2 \leq \max\{b_i\}/n \leq \max\{n_i\}/n$. It is worth noting that, to recover the underlying data sampled from UoS, it is equivalent to find a matrix $L$ and $W$ under the above constraints. The space of $W$ can be then defined as follows, 
\newtheorem{mydef}{Definition}

\begin{mydef}
\emph{(k-block-diagonal matrix)} We say that a $n\times n$ matrix $M$ is k-block-diagonal if and only if
\begin{enumerate}
	\item There exists a permutation matrix $P$, such that $\tilde{M} = PMP^{-1}$ is a block-diagonal matrix
	\item the maximum dimension of each block of $\tilde{M}$ is less or equal than $k+1$.
\end{enumerate}
The space of all such matrices is denoted as $BM_k$. 
\end{mydef}
We next define the space of $L$ based on the space $BM_k$ of $W$.

\begin{mydef}
\emph{(k-self-representative matrix)}. We say that a $d\times n$ matrix $X$ with no zero column is k-self-representative if and only if
$$
X = XW, W \in BM_k, W_{ii} = 0.
$$
The space of all such $d\times n$ matrices is denoted by $SR_k$
\end{mydef}

Consider the case that sample $l_i$ is corrupted by some sparse error $e_i$. Intuitively, we want to separate the sparse errors from the data matrix $X$ and present the remainder in $SR_k$. Therefore Problem \ref{prob1} can be formulated as
\begin{align}
\label{origProb}
&\min \|E\|_0 \\
&s.t. X = L + E, L \in SR_k.	\notag
\end{align}
There are some fundamental difficulties in solving this problem on account of the combinatorial nature of $\|\cdot\|_0$ and the complicated geometry of $SR_k$. The results of an $l_1$ norm approximation is useful to establish the sparsity of $E$ \cite{wright10}. The real difficulty, however, is that not only $SR_k$ is a non-convex space,\footnote{Consider $M_1, M_2 \in SR_k$, let $M_1 = \left( \begin{array}{cc}
1 & 2 \\ 0 & 0
\end{array} \right)$ and 
$M_2 = \left( \begin{array}{cc}
0 & 0 \\ 2 & 1
\end{array} \right)$. It is easy to see that $M = (M_1 + M_2)/2 = \left( \begin{array}{cc} 1 & 2 \\ 2 & 1
\end{array} \right) \notin SR_2$. }
and even worse, $SR_k$ is not path-connected. Intuitively, it is helpful to consider $L_1, L_2 \in SR_k$, and let $col(L_1) \cap col(L_2) = 0$, then all possible paths connecting $L_1$ and $L_2$ must pass the origin, given that $L$ is a matrix with no zero columns, and $0 \notin SR_k$. $SR_k$ can hence be divided into at least two components $S_p$ and $SR_k/S_p$. 

To avoid solving (\ref{origProb}) with a disconnected feasible region, we opt to integrate this constraint into the objective function, and see the problem from a different angle. We hence have the following definition:

\begin{mydef}
\emph{($\mathcal{W}_0$-function on a matrix space).} For any $d\times n$ matrix $X$, if there is $W\in BM_k$, such that $X = XW$, then
\begin{align*}
\mathcal{W}_0(X) = \min_{W} \|W\|_0,
~~s.t.~ X = XW, W_{ii} = 0, W \in BM_k.
\end{align*}
Otherwise, $\mathcal{W}_0(X) = \infty$
\end{mydef}
Then instead of (\ref{origProb}), we consider the following optimization problem:
\begin{align}
\label{eqn1}
&\min_{L,E} \mathcal{W}_0(L) + \lambda\|E\|_0 \\
&s.t. X = L + E.	\notag
\end{align}
The relation of (\ref{origProb}) and (\ref{eqn1}) is established by the following lemma:
\newtheorem{mylemma}{Lemma}
\begin{mylemma}
\label{lemma1}
For certain $\lambda$, if $(\hat{L}, \hat{E})$ is a pair of global optimizer of (\ref{eqn1}), then $(\hat{L}, \hat{E})$ is also a global optimizer of (\ref{origProb}). 
\end{mylemma}
The proof of Lemma \ref{lemma1} is presented in Appendix \ref{appA}.  

Next we will leverage the parsimonious property of $l_1$ norm to approximate $\|\cdot\|_0$. First, the definition of $\mathcal{W}_0(\cdot)$ is extended to a $l_1$ norm-based function:
\begin{mydef}
\emph{($\mathcal{W}_1$-function on a matrix space).} For any $d\times n$ matrix $X$, if there exists $W\in BM_k$, such that $X = XW$, then
\begin{align*}
\mathcal{W}_1(X) = \min_{W} \|W\|_1,
~~s.t.~ X = XW, W_{ii} = 0, W \in BM_k.
\end{align*}
Otherwise, $\mathcal{W}_1(X) = \infty$
\end{mydef}

We then have the following problem, 
\begin{align}
\label{eqn2}
&\min \mathcal{W}_1(L) + \lambda\|E\|_1 \\
&s.t. X = L + E \notag
\end{align}

It is worth noting that formulation (\ref{eqn2}) bears a similar form to the problem of robust PCA in \cite{RPCA}. Intuitively, both problems attempt to decompose the data matrix into two parts: one with a parsimonious support, and the other also with a sparse support, however in a different domain. For robust PCA, the parsimonious support of the low rank matrix lies in the singular values. In our case, the sparse support of $L$ lies in the matrix $W$ in the $\mathcal{W}_0$ function, meaning that columns of $L$ can be sparsely self-represented.

\section{Recovery of a union of subspaces: conditions and methodologies}
\label{SRmethod}
\subsection{A sufficient condition for exact recovery}
\label{suffcond}

In this section, we discuss the important question of when the underlying structure can be exactly recovered by solving (\ref{eqn2}). This problem is essentially twofold: first, it is about when the solution of $(\hat{L},\hat{E})$ is exact; and second, when $\hat{W}$ correctly reflects the true clustering structure. For the former, we establish a sufficient condition of exact decomposition of $L$ and $E$ as follows:
\newtheorem{mythm}{Theorem}
\begin{mythm}
\label{thm1}
$(L_0,E_0)$ can be exactly recovered by solving (\ref{eqn2}) with $\lambda > 0$, $i.e. (\hat{L},\hat{E}) = (L_0, E_0)$, if $\forall A$ for same dimension of $L$, at least one of the following conditions is true:
\begin{enumerate}
\item for any partition of $L_0 = [L_1 | L_2 | \dots | L_J], ~|col(L_I)|<k+1$, and $A = [A_1 | A_2 | \dots | A_J]$ accordingly, $\exists I$, such that $L_I + A_I$ is full rank.
\item $\|P_{\Omega_{E}^c}A\|_1 - \|P_{\Omega_{E}}A\|_1 \geq \frac{\|W_0\|_1}{\lambda}$.
\end{enumerate}
\end{mythm}
The proof of Theorem \ref{thm1} is presented in Appendix \ref{appthm}. Specifically, the first condition means that the perturbation $A$ on $L$ could lead to a non-feasible point, and the second condition states that $E$ is sparse in a way that any feasible move will create a larger component outside the support of $E$ then inside. Intuitively, this theorem states that the space $SR_k$ and $E$ should be nearly "incoherent" to each other, in the sense that any change of $L_0$, $i.e.~A = L'-L_0$, will make $E' = E_0 - A$ less sparse, and on the other hand, any sparse solution $E'$ will move the corresponding $L'$ off of space $SR_k$. 

After having exact $L$ and $E$, the problem of finding $W$ of $\mathcal{W}_1$ given $L$ is equivalent to subspace clustering without outliers, and therefore the correctness is substantiated by Theorem 2.5 in \cite{sol2012}. Concretely, this theorem guarantees that if the underlying subspaces are not too "close", and the distribution of points in each subspace is not too skewed, then $w_{ij} \neq 0$ if and only if $l_i$ and $l_j$ are in the same subspace. 

\subsection{An approximate solution via sparse modeling}
Under the conditions stated in Section \ref{suffcond}, also substantiated by \cite{sol2012}, finding $\mathcal{W}_1(L)$ can be accomplished by tunning the condition $L\in SR_k$ to $W_{ii} = 0$, subsequently modifying $\mathcal{W}_1(L)$ into a convex function and making it defined in a connected domain. Intuitively, since the sparsity of any $W \in BM_k$ is upper bounded by $\max\{b_i\}/n$, where $b_i$ is the size of the maximum block, we are essentially looking for a sparse $W$. Furthermore, the constraint of $W\in BM_k$ would be unnecessary when a sparse feasible solution has the same block-diagonal structure reflecting the structure of a union of subspaces. It is exactly the case when any pair of subspaces for the given dataset are not too "close" to each other. More thorough analysis can be found in \cite{sol2012}. 

We therefore have
\begin{align}
\mathcal{\tilde{W}}_1(L) = \min_{W} \|W\|_1,
~~s.t.~ L = LW, W_{ii} = 0. 
\end{align}

Substituting $\mathcal{W}_1(L)$ by $\mathcal{\tilde{W}}_1(L)$ in (\ref{eqn2}), it allows us to relax the constraints of (\ref{eqn2}) and directly work on the following problem,   

\begin{align}
\label{mainEqn}
&\min_{W,E} \|W\|_1 + \lambda\|E\|_1, \\
&s.t. X = L + E, L = LW, W_{ii} = 0.	\notag
\end{align}

Other than posing this problem as a recovery and clustering problem, we may also view it from a dictionary learning angle. Note that the constraint $X = L + E$ may be rewritten as $X = LW + E$, to therefore reinterpret the problem of finding $L$ and $E$ as a dictionary learning problem. In addition to the sparse model, atoms in dictionary $L$ are achieved from data samples with sparse variation. It hence may be seen as a generalization of \cite{elham2012see} in the sense that we not only pick representative samples from the given data set using $l_1$ norm, but also adapt the representative samples so that they can "fix" themselves and hence be robust to sparse errors. 

\section{Algorithm: Subspace Recovery via Bi-Sparsity Pursuit}
\label{algorithm}
Obtaining an algorithmic solution to (\ref{mainEqn}) is complicated by the bilinear term in constraints which lead to a non-convex optimization. In this section, we leverage the successes of alternating direction method (ADM)~\cite{lin2010augmented} and linearized ADM (LADM)~\cite{lin2011linearized} in large scale sparse representation problem, and focus on designing an appropriate algorithm to approximate the global minimum of (\ref{mainEqn}). 

Our method, what we refer to as robust subspace recovery via bi-sparsity pursuit (RoSuRe), is based on linearized ADMM~\cite{lin2011linearized}. Concretely, we pursue the sparsity of $E$ and $W$ alternatively until convergence. Besides the effectiveness of ADMM on $l_1$ minimization problems, a more profound rationale for this approach is that the augmented Lagrange multiplier (ALM) method can address the non-convexity of (\ref{mainEqn})~\cite{rock1974augmented}\cite{lu2003linear}. Specifically, Theorem 4 in ~\cite{rock1974augmented} states that under the ALM setting, when a solution exists and the objective function is lower bounded, the duality gap is zero. It hence follows that with a sufficiently large augmented Lagrange multiplier $\mu$, we can approximate the global optimizer by solving the dual problem. 

\begin{algorithm}
\label{alg1}
\caption{Subspace Recovery via Bi-Sparsity Pursuit (RoSuRe)}
\begin{algorithmic}
\item Initialize: Data matrix $X \in R^{m \times n}$, $\lambda$, $\rho$, $\eta_1$, $\eta_2$

\While{not converged}
	\item Update $W$ by linearized soft-thresholding
	\State $L_{k+1} = X - E_{k}$, 
	\State $W_{k+1} = \mathcal{T}_{\frac{1}{\mu\eta_1}}\left(W_k + \frac{L_{k+1}^T(L_{k+1}\hat{W}_{k} - Y_{k}/\mu_k)}{\eta_1}\right)$. 
	\State $W_{k+1}^{ii} = 0$. 

	\item Update $E$ by linearized soft-thresholding 
	\State $\hat{W}_{k+1} = I - W_{k}$,
	\State $E_{k+1} = \mathcal{T}_{\frac{1}{\mu\eta_2}}\left(E_k + \frac{(L_{k+1}\hat{W}_{k+1} - Y_k/\mu_k)\hat{W}_{k+1}^T}{\eta_2}\right)$

\item Update the lagrange multiplier $Y$ and the augmented lagrange multiplier $\mu$
	\State $Y_{k+1} = Y_{k} + \mu_k(L_{k+1}W_{k+1}-L_{k+1})$
	\State $\mu_{k+1} = \rho\mu_k$

\EndWhile
\end{algorithmic}
\end{algorithm}

Specifically, substituting $L$ by $X-E$, and using $L = LW$, we can reduce (\ref{mainEqn}) to a two-variable problem, and hence write the augmented Lagrange function of (\ref{mainEqn}) as follows, 
\begin{align}
&L(E,W,Y,\mu) \notag \\
= &\lambda\|E\|_1 + \|W\|_1 + \langle (X-E)W - (X-E), Y \rangle \notag \\
+ &\frac{\mu}{2}\|(X-E)W-(X-E)\|_F^2.
\end{align}
Letting $\hat{W} = I - W$, we alternatively update $W$ and $E$, 
\begin{align}
\label{wp}
W_{k+1} &= \arg\min_{W} \|W\|_1 + \langle L_{k+1}W-L_{k+1}, Y_k\rangle\notag \\
&+ \frac{\mu}{2}\|L_{k+1}W-L_{k+1}\|_F^{2}, \\ 
\label{ep}
E_{k+1} &= \arg\min_{E} \lambda\|E\|_1 + \langle -L_{k+1} \hat{W}_{k+1}, Y_k \rangle \notag\\
&+ \frac{\mu}{2}\|L_{k+1} \hat{W}_{k+1}\|_F^2. 
\end{align}
The solution of (\ref{wp}) and (\ref{ep}) can be well approximated in each iteration by linearizing the augmented Lagrange term ~\cite{lin2011linearized}, 
\begin{align}
W_{k+1} &= \mathcal{T}_{\frac{1}{\mu\eta_1}}\left(W_k + \frac{L_{k+1}^T(L_{k+1}\hat{W}_{k} - Y_{k}/\mu_k)}{\eta_1}\right), \\
E_{k+1} &= \mathcal{T}_{\frac{1}{\mu\eta_2}}\left(E_k + \frac{(L_{k+1}\hat{W}_{k+1} - Y_k/\mu_k)\hat{W}_{k+1}^T}{\eta_2}\right),
\end{align}
where $\eta_1 \geq \|L\|^2_2$, $\eta_2 \geq \|\hat{W}\|^2_2$, and $\mathcal{T}_\alpha(\cdot)$ is a soft-thresholding operator. 

In addition, the Lagrange multipliers are updated as follows,
\begin{align}
Y_{k+1} &= Y_{k} + \mu_k(L_{k+1}W_{k+1}-L_{k+1}) \\
\mu_{k+1} &= \rho\mu_k
\end{align}

\section{Experiments and Validations}
\label{experiments}

\subsection{Experiments on Synthetic Data}

Section \ref{SRmethod} discusses the sufficient condition to recover data structure by solving (\ref{origProb}). In this section, we hence empirically investigate the viability extent of RoSuRe with various conditions. The recovery results are compared with Robust PCA~\cite{RPCA} using the method presented in~\cite{lin2010augmented} and sparse subspace clustering using the algorithm in~\cite{elham2013sparse}. 

\begin{figure}[h]
\begin{center}
\begin{tabular}{ccc}

\includegraphics[width=0.25\textwidth]{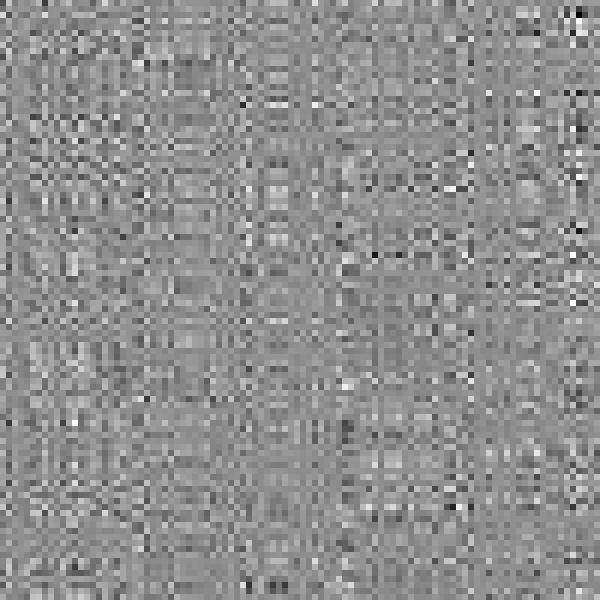} &
\includegraphics[width=0.25\textwidth]{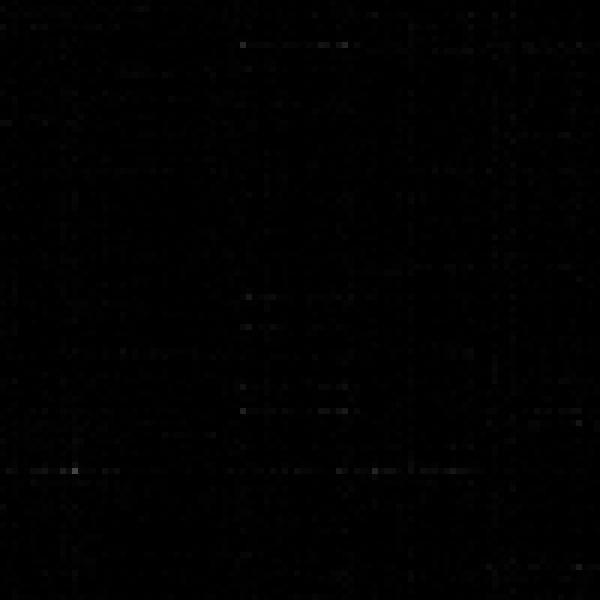} & 
\includegraphics[width=0.25\textwidth]{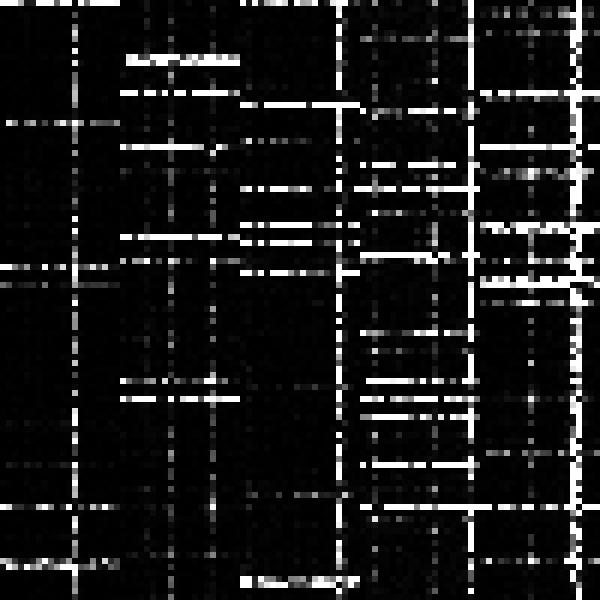} \\
\small (a)$L_0$ & \small (b)$|L_0-L_{RoSuRe}|$ & \small (c)$|L_0 - L_{RPCA}|$ \\

\includegraphics[width=0.25\textwidth]{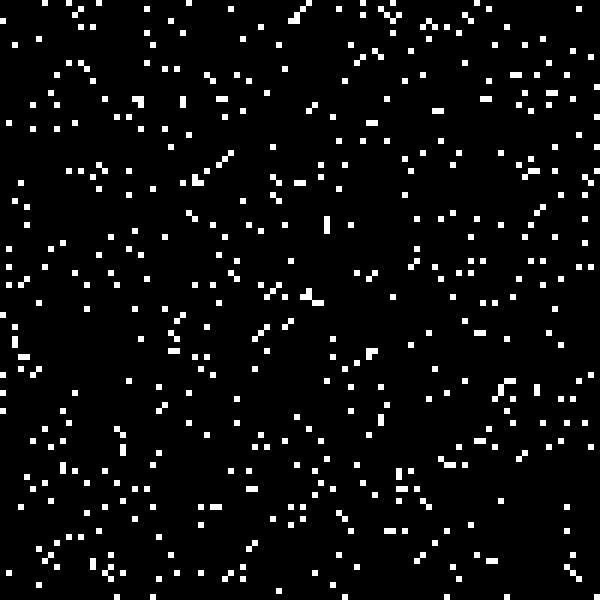} &
\includegraphics[width=0.25\textwidth]{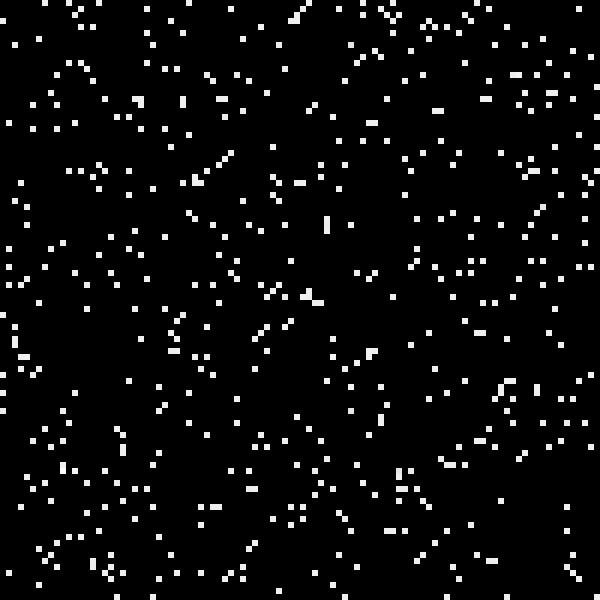} &
\includegraphics[width=0.25\textwidth]{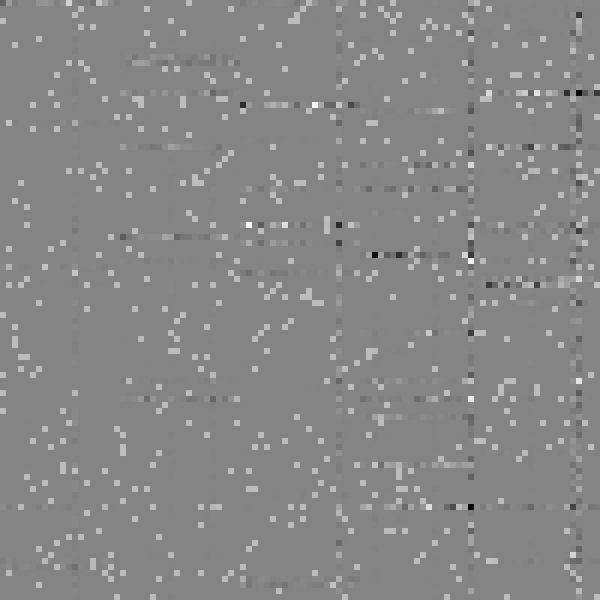} \\
\small (d)$E_0$ & \small (e)$E_{RoSuRe}$ & \small (f)$E_{RPCA}$ \\

\end{tabular}
\end{center}
\caption{An example of subspace exact recovery and comparison with robust PCA}
\label{fig:rsrFig}
\end{figure}

\begin{figure}[h]
\begin{center}
\includegraphics[width=0.30\textwidth]{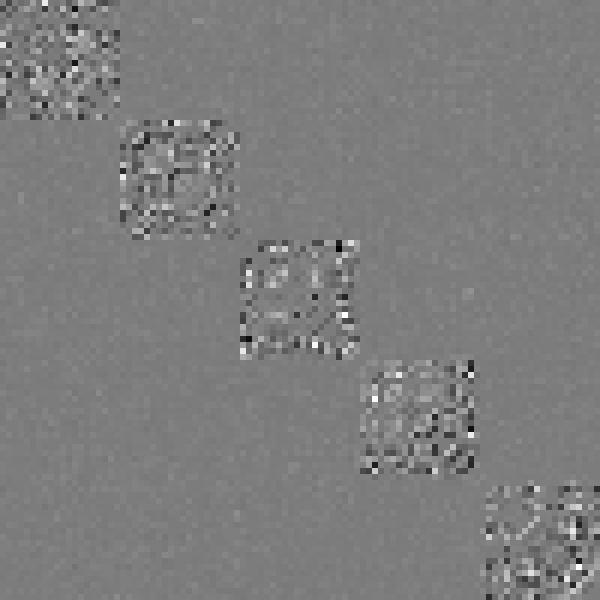} \\

\end{center}
\caption{The sparse coeffcieint matrix $W$}
\label{rsrfig2}
\end{figure}

The data matrix is fixed to be a $200\times200$ matrix, and all data points are sampled from a union of 5 subspaces. Fig.\ref{fig:rsrFig} shows the results of an example of the exact recovery and the comparison with robust PCA. Note that $(L_{RoSuRe}, E_{RoSuRe})$ and $(L_0,E_0)$ are almost identical. In Fig.\ref{rsrfig2}, we can see that $W_{RoSuRe}$ shows clear clustering properties such that $w_{ij} \approx 0$ when $l_i,l_j$ are not in the same subspace.

\begin{figure}
\begin{center}
\begin{tabular}{ccc}
\includegraphics[width = 0.29\textwidth]{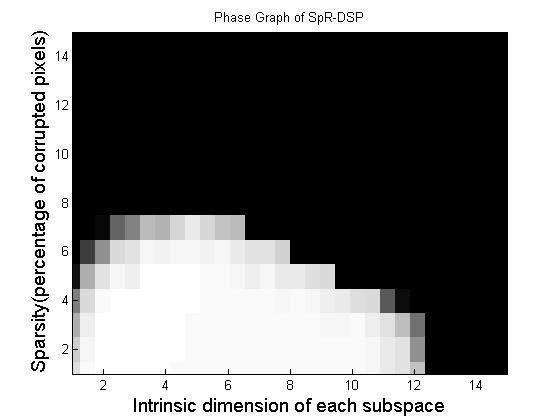} &
\includegraphics[width = 0.29\textwidth]{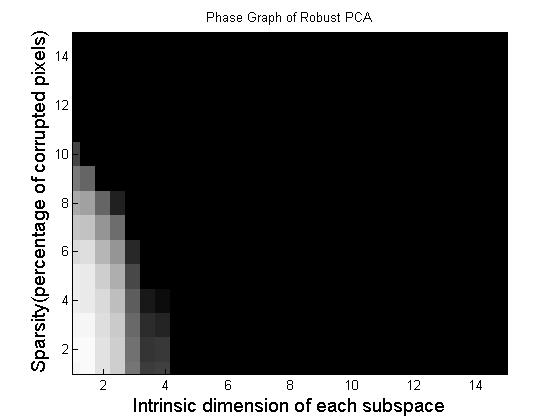} &
\includegraphics[width = 0.29\textwidth]{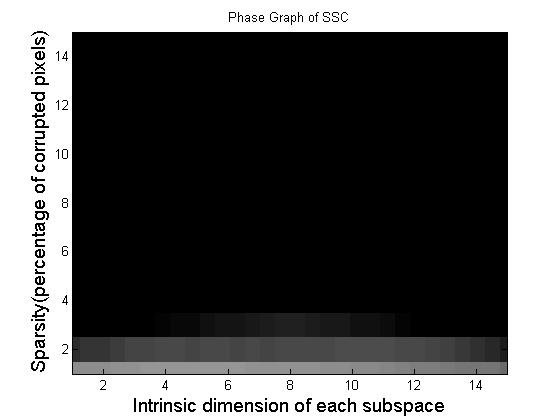} \\
\small(a)RoSuRe & \small(b)Robust PCA & \small(c) SSC \\
\end{tabular}
\end{center}
\caption{Overall recovery results of RoSuRe and Robust PCA. [0 0.2] is mapped to [1 0] of grayscale image}
\label{pt}
\end{figure}

Fig.\ref{pt} is the overall recovery results of RoSuRe, robust PCA and SSC. White shaded area means a lower error and hence amounts to exact recovery. The dimension of each subspace is varied from 1 to 15, and the sparsity of $S$ from 0.5\% to 15\%. Each submatrix $L_I = X_IY_I^T$ with $n\times d$ matrices $X_I$ and $Y_I$, are independently sampled from an i.i.d normal distribution. The recovery error is measured as $err(L) = \|L_0 - \hat{L}\|_F/\|L_0\|_F$. We can see a significant larger range of RoSuRe compared to robust PCA and SSC. The reason to the result of RoSuRe and robust PCA is due the difference of data models. Concretely, when the sum of the dimension of each subspace is small, the UoS model degenerates to a "low-rank + sparse" model, which suits robust PCA very well. On the other hand, when the dimension of each subspace increases, the overall rank of $L$ tend to be accordingly larger and hence the low rank model may not hold anymore. Since RoSuRe is designed to fit UoS model, it can recover the data structure in a wider range. For SSC, this method specifically fit the condition when only a small portion of data are outliers. Under the assumption that most of the data is corrupted, it is hence very difficult to reconstruct samples by other corrupted ones. 

\subsection{Experiments on Computer Vision Problems}
Since UoS model has been intensively researched and successfully applied to many computer vision and machine learning problems~\cite{liu2013}~\cite{elham2013sparse}~\cite{elad2012sparse}, we expect that our model may also fit these problems. Here, we present experimental results of our method on video background subtraction and face clustering problem, as exemplars of the promising potential. 

\subsubsection{Video background subtraction}
\label{vbsub}
\begin{figure}[h]
\begin{center}
\begin{tabular}{ccc}
\includegraphics[width = 0.20\textwidth]{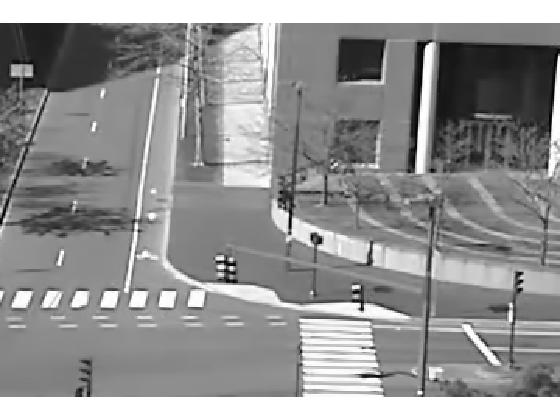} &
\includegraphics[width = 0.20\textwidth]{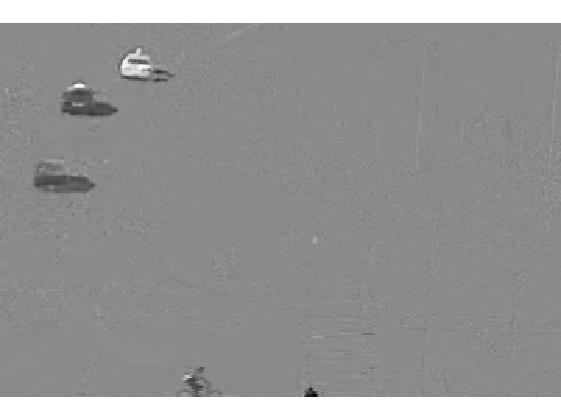} &
\includegraphics[width = 0.20\textwidth]{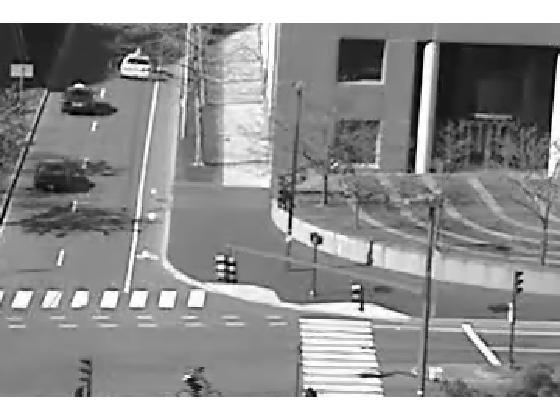} \\

\includegraphics[width = 0.20\textwidth]{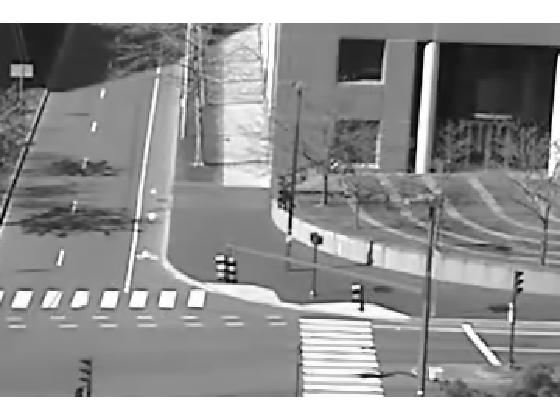} &
\includegraphics[width = 0.20\textwidth]{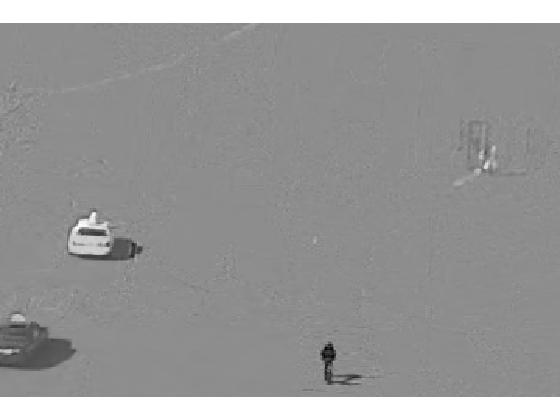} &
\includegraphics[width = 0.20\textwidth]{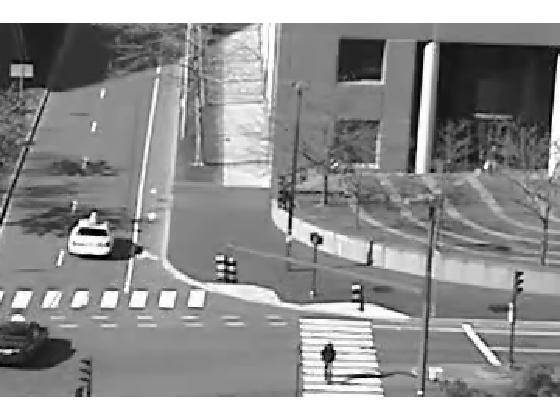} \\

\small(a)Background & \small(b)Foreground & \small(c)Original frame\\

\end{tabular}
\end{center}
\caption{Background subtraction on traffic videos (static camera)}
\label{staticVideo}
\end{figure}

\begin{figure}[h]
\begin{center}
\begin{tabular}{ccc}
\includegraphics[width = 0.20\textwidth]{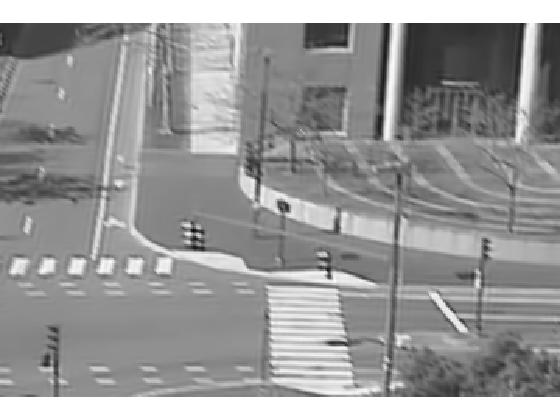} &
\includegraphics[width = 0.20\textwidth]{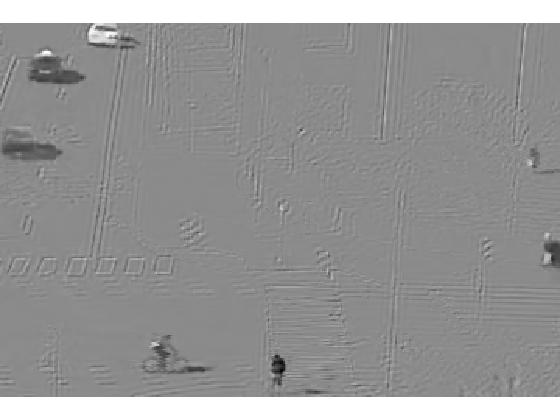} &
\includegraphics[width = 0.20\textwidth]{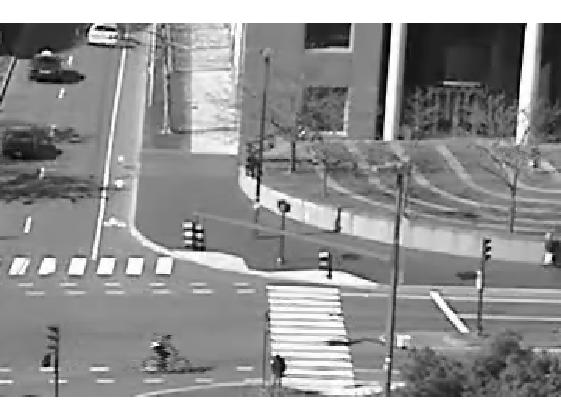} \\

\includegraphics[width = 0.20\textwidth]{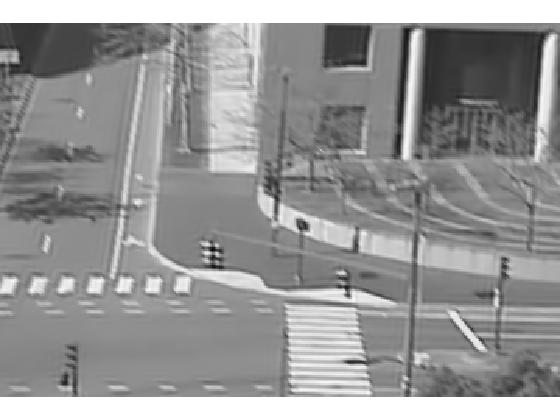} &
\includegraphics[width = 0.20\textwidth]{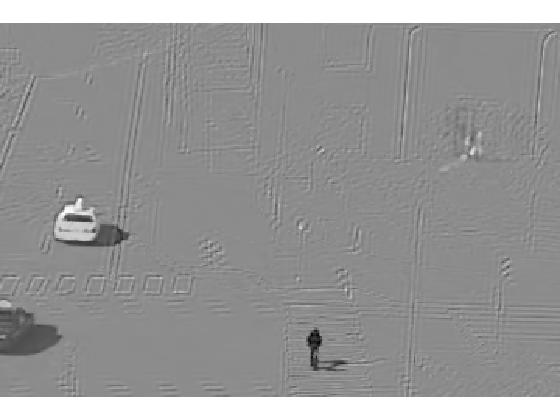} &
\includegraphics[width = 0.20\textwidth]{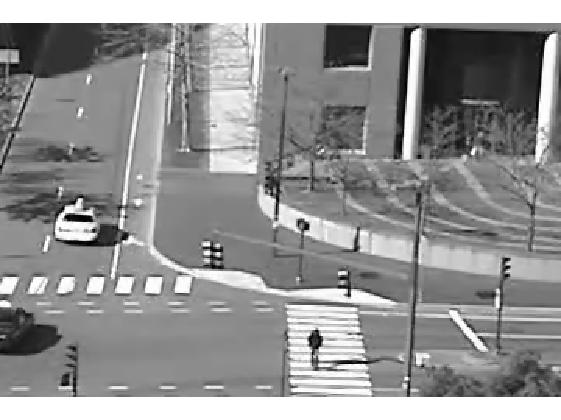} \\

\small(a)Background & \small(b)Foreground & \small(c)Original frame\\

\end{tabular}
\end{center}
\caption{Background subtraction with panning camera}
\label{movingVideo}
\end{figure}

Surveillance videos can be naturally modeled as UoS model due to their relatively static background and sparse foreground. The power of our proposed UoS model lies in coping with both a static camera and a panning one with periodic motion. Here we test our method in both scenarios using surveillance videos from MIT traffic dataset~\cite{wang2009}. In Fig.\ref{staticVideo}, we show the segmentation results with a static background. For the scenario of a "panning camera", we generate a sequence by cropping the previous video. The cropped region is swept from bottom right to top left and then backward periodically, at the speed of 5 pixels per frame. The results are shown in Fig.\ref{movingVideo}. We can see that the results in the moving camera scenario are only slightly worse than the static case. 

\begin{figure}[h]
\begin{center}
\begin{tabular}{cc}
\includegraphics[width = 0.30\textwidth]{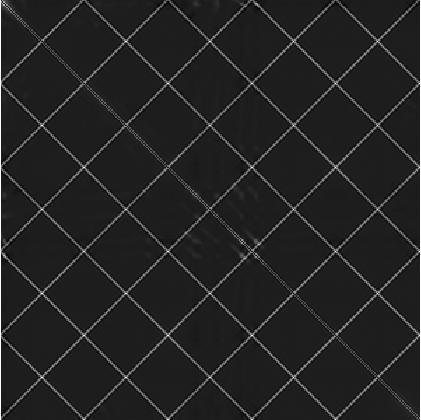} &
\includegraphics[width = 0.30\textwidth]{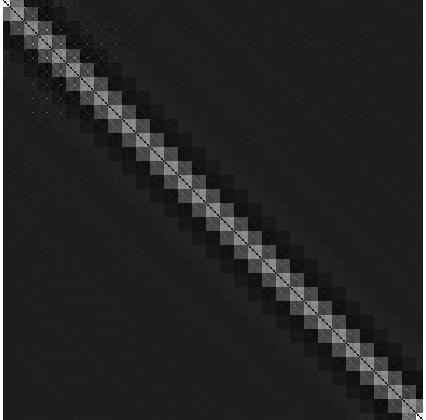} \\
(a) & (b) \\
\end{tabular}
\end{center}
\caption{Coefficient matrix $W$ (a) without rearrangement according to the position of the camera (b) with rearrangement according to the position of the camera}
\label{tfW}
\end{figure}

More interestingly, the sparse coefficient matrix $W$ provides important information about the relations among data points, which potentially may be used to cluster data into individual clusters. In Fig. \ref{tfW}(a), we can see that, for each column of the coefficient matrix $W$, the nonzero entries appear periodically. In considering the periodic motion of the camera, we essentially mean that every frame is mainly represented by the frames when the camera is in a similar position, i.e. a similar background, with the foreground moving objects as outliers. We hence permute the rows and columns of $W$ according to the position of cameras, as shown in Fig. \ref{tfW}(b). A block-diagonal structure then emerges, where images with similar backgrounds are clustered as one subspace. 

\subsubsection{Face clustering under various illumination conditions}

Recent research on sparse models implies that a parsimonious representation may be a key factor for classification~\cite{elad2012sparse}~\cite{lee2007sparse}. Indeed, the sparse coefficients pursued by our method shows clustering features in experiments of both synthetic and real-world data. To further explore the ability of our method, we evaluate the clustering performance on the Extended Yale face database B~\cite{KCLee05}, and compare our results to those of state-of-the-art methods~\cite{yan2006general}~\cite{liu2013}~\cite{elham2013sparse}. 

\begin{figure}[h]
\begin{center}
\includegraphics[width = 0.75\textwidth]{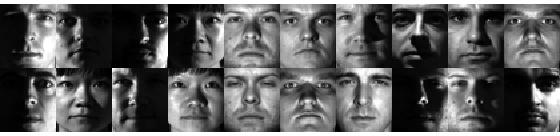}
\end{center}
\caption{Sample face images in Extended Yale face database B}
\label{faceSample}
\end{figure}

The database includes cropped face images of 38 different people under various illumination conditions. Images of each person may be seen as data points from one subspace, albeit heavily corrupted by entries due to different illumination conditions, as shown in Fig.~\ref{faceSample}. In our experiment, we adopt the same setting as \cite{elham2013sparse}, such that each image is downsampled to $48\times42$ and is vectorized to a 2016-dimensional vector. In addition, we use the sparse coefficient matrix $W$ from RoSuRe to formulate an affinity matrix as $A = \tilde{W} + \tilde{W}^T$, where $\tilde{W}$ is a thresholded version of $W$. The spectral clustering method in \cite{ng2002spectral} is utilized to determine the clusters of data, with affinity matrix $A$ as the input. 

\begin{table}
	\caption{Clustering error (\%) on the Extended Yale Face Database B compared to state-of-the-art methods~\cite{elham2013sparse}~\cite{liu2013}~\cite{yan2006general}}
	\begin{center}
	\begin{tabular} {| c | c | c | c | c |}
	\hline
	Algorithm & LSA & LRR & SSC & RoSuRe \\
	\hline
	\hline
	2-subjects Mean& 38.20 & 2.54 & 1.86 & \textbf{0.71} \\
	Median & 47.66 & 0.78 & \textbf{0.00} & 0.39 \\
	\hline
	\hline
	5-subjects Mean& 58.02 & 6.90 & 4.31 & \textbf{3.24} \\
	Median & 56.87 & 5.63 & 2.50 & \textbf{1.72} \\
	\hline
	\hline
	10-subjects mean& 60.42 & 22.92 & 10.94 & \textbf{5.62} \\
	Median & 57.50 & 23.59 & 5.63 & \textbf{5.47} \\
	\hline
	\end{tabular}
	\end{center}
	\label{table1}
\end{table}

\begin{figure}[H]
\begin{center}
\includegraphics[width = 0.6\textwidth]{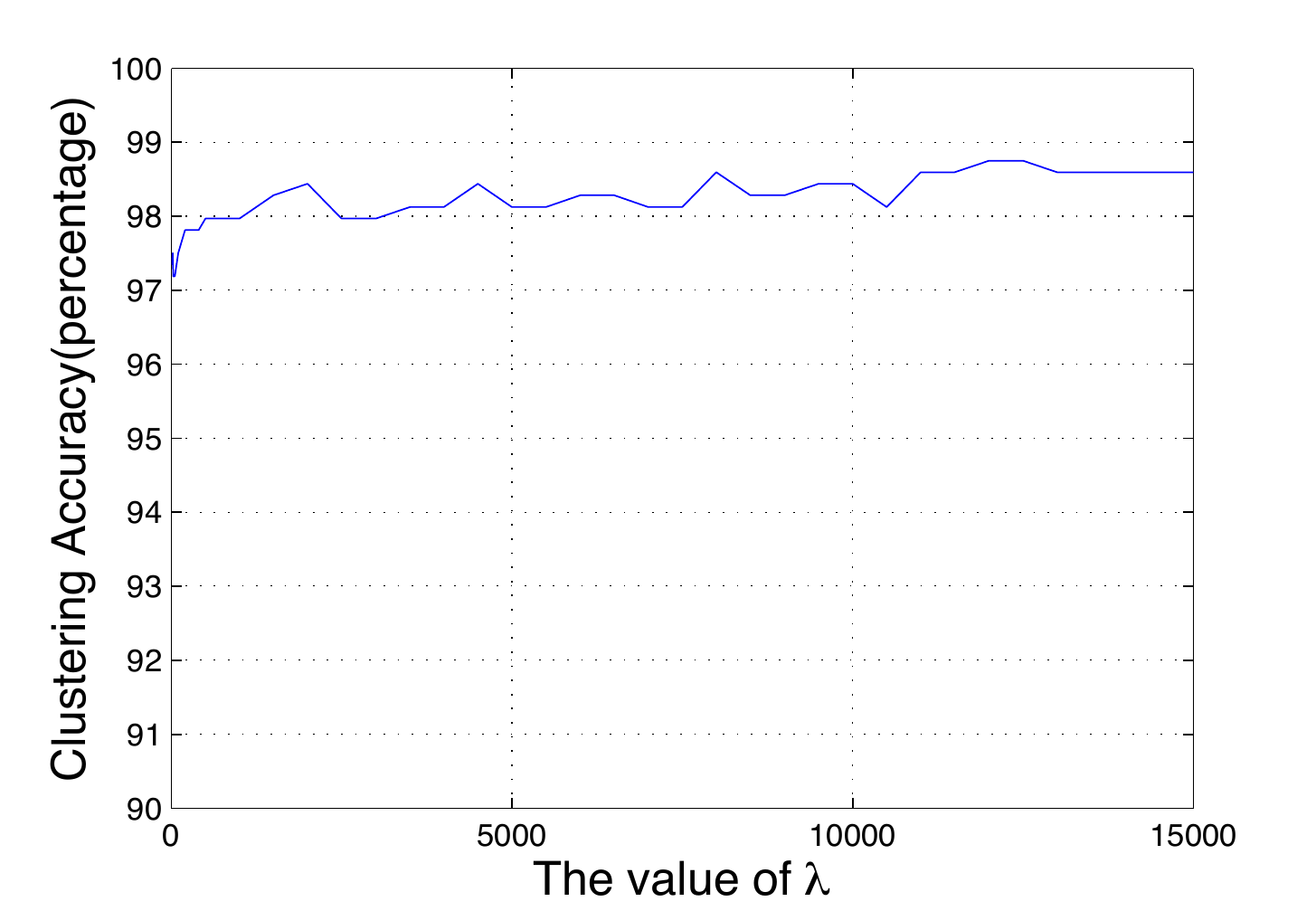}
\end{center}
\caption{Clustering Accuracy vs The value of $\lambda$ in face clustering of 10 subjects}
\label{clusterAccuracy}
\end{figure}

\begin{figure}[h]
\begin{center}
\includegraphics[width = 0.7\textwidth]{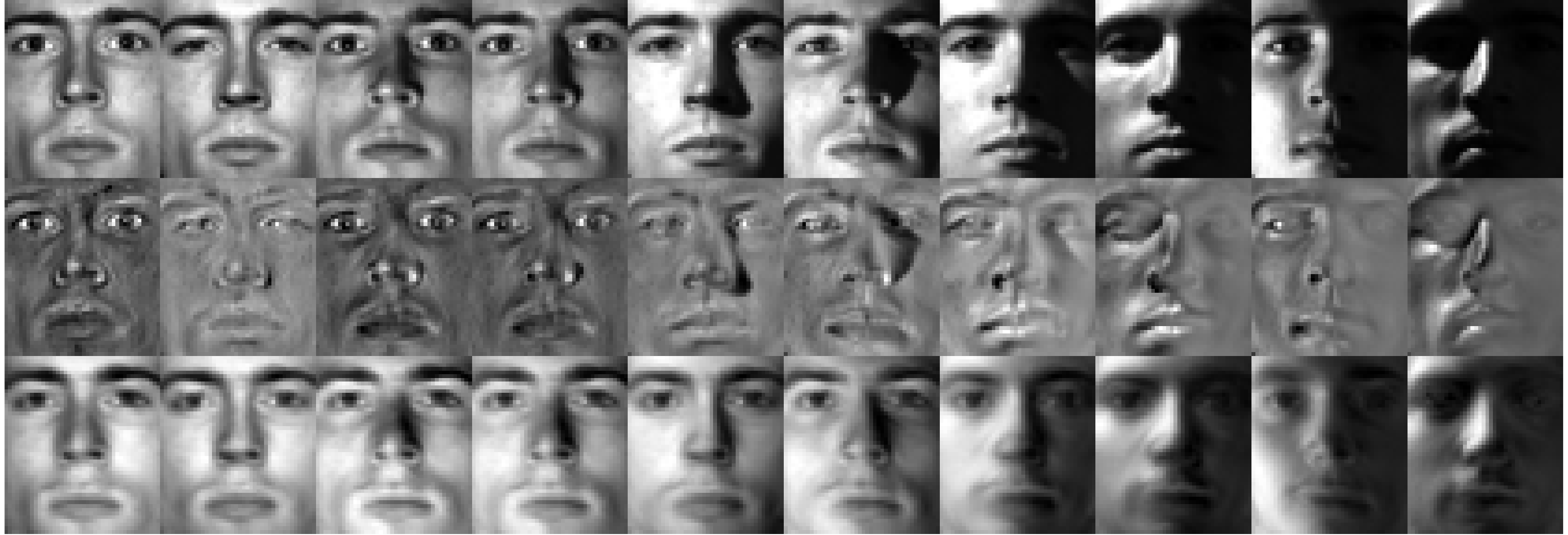}
\end{center}
\caption{Recovery results of human face images. The three rows from top to bottom are original images, the components $E$, and the recovered images, respectively. }
\label{faceRecovery}
\end{figure}

We compare the clustering performance of RoSuRe with the state-of-the-art methods such as local subspace analysis(LSA)~\cite{yan2006general}, sparse subspace clustering (SSC)~\cite{elham2013sparse}, and low rank representation(LRR)~\cite{liu2013}. The best performance of each method is referenced in Table \ref{table1} for comparison. As shown in the table, RoSuRe has the lowest mean clustering error rate in all three settings, i.e. 2 subjects, 5 subjects and 10 subjects. In particular, in the most challenging case of 10 subjects, the mean clustering error rate is as low as 5.62\% with the median 5.47\%. In addition, we show the robustness of our method with respect to $\lambda$ in a 10-subject scenario. In Fig. \ref{clusterAccuracy}, the cluster accuracy maintains above 98\% with $\lambda$ varying from 500 to 15000. 

In Fig. \ref{faceRecovery}, we present the recovery results of some sample faces from the 10-subject clustering scenario. In most cases, the sparse term $E$ compensates the information missing caused by lightning condition. This is especially true when the shadow area is small, i.e. a sparser support of error term $E$, we can see a visually perfect recovery of the missing area. This result validates the effectiveness of our method to solve the problem of subspace clustering with sparsely corrupted data. 

\section{Conclusion}
We have proposed in this paper a novel approach to recover underlying subspaces of data samples from measured data corrupted by general sparse errors. We formulated the problem as a non-convex optimization problem, and a sufficient condition of exact recovery is proved. We also designed an effective algorithm named RoSuRe to well approximate the global solution of the optimization problem. Furthermore, experiments on both synthetic data and real-world vision data are presented to show a broad range of applications of our method.  

Future work may include several aspects across computer vision and machine learning. It would first be interesting to understand and extend this work from a dictionary learning angle, to learn a feature set for high dimensional data representation and recognition. Additionally, the sufficient condition proved in this paper is fairly strong, and a weaker condition is not only theoretically interesting, but would also be helpful for better understanding the problem. 

\section{Acknowledgement}
Thanks to the gracious support of the Missile Defense Agency (MDA) under HQ0147-11-C-6012 contract.

\appendix
\section{Proof of Lemma \ref{lemma1}}
\label{appA}

At the beginning, we rewrite the objective function in (\ref{eqn1}) as 
\begin{align}
f(L,E) = \frac{\mathcal{W}_0(L)}{\lambda} + \|E\|_0.
\end{align}
It is clear that this will not change the minimum value. In addition, we assume that there exists $L \in SR_k$, otherwise the statement would be trivial, since (\ref{origProb}) would be not feasible, and the value of the objective function in (\ref{eqn1}) would be infinite.  

Let $(\hat{L}, \hat{E})$ be a global minimizer of (\ref{eqn1}), then $\hat{L}\in SR_k$. If $\exists~E'$, such that $\|E'\|_0 < \|\hat{E}\|_0$ and $L'=X-E'\in SR_k$,  we have
\begin{align}
\label{p1step1}
f(L',E') &= \|E'\|_0 + 1 + \frac{\mathcal{W}_0(L')}{\lambda} - 1 \\ \notag
		   &\leq \|\hat{E}\|_0 + \frac{\mathcal{W}_0(L')}{\lambda} - 1. 
\end{align}
Since $(\hat{L}, \hat{E})$ is a global minimizer, $f_2(\hat{L},\hat{E}) < f_2(L',E')$. Combined with (\ref{p1step1}), 
\begin{align}
0 < f(L',E') - f(\hat{L},\hat{E}) \leq \frac{\mathcal{W}_0(L')-\mathcal{W}_0(\hat{L})}{\lambda} - 1. 
\end{align}
Then it follows that
\begin{align}
\label{p1step2}
\lambda < \mathcal{W}_0(L')-\mathcal{W}_0(\hat{L}).
\end{align}
Note that when $L\in SR_k$, $0< \mathcal{W}_0(L) \leq n^2$, where $n$ is the number of columns of $L$. Therefore, letting $\lambda \geq n^2$ will violate (\ref{p1step2}) since
\begin{align}
\lambda \geq n^2 > \mathcal{W}_0(L') - \mathcal{W}_0(\hat{L}). 
\end{align}
Hence, with $\lambda \geq n^2$, $\hat{E}$ is also a solution of (\ref{origProb}). Lemma \ref{lemma1} is proved. 

\section{Proof of Theorem \ref{thm1}}
\label{appthm}
$\forall A$ such that $dim(A) = dim(L_0)$, we first prove by contradiction that for any partition of $L_0 = [L_1 | L_2 | \dots | L_J], col(L_I)\leq k+1$, if there exists $I$, such that $L_I+A_I$ is full rank, then $(L',E') = (L_0+A,E_0-A)$ is not feasible. 

Assume that $L' = L_0 + A$ is feasible, then there is a $W'\in BM_k$ such that $L' = L'W'$. Partition $L'$ according to the block form of $W'$ as 
$$L' = [L'_1 | L'_2 | \dots | L'_{J'}], col(L'_I)\leq k+1,$$
then for each $I$, we have $ L'_I = L'_IW'_I$ and $L'_I$ is not full-rank, which contradicts the assumption that $\exists~I$ such that $L_I+A_I$ is full-rank. 

Next we prove that if $\|P_{\Omega_{E}}^cA\|_1 - \|P_{\Omega_{E}}A\|_1 \geq \sigma$, then for $(L',E') = (L_0+A,E_0-A)$, $f(L,E) < f(L',E')$. 
Consider
\begin{align}
&f(L',E') - f(L,E) \notag \\
&= \|E_0-A\|_1 - \|E_0\|_1 + \frac{\|W'\|_1}{\lambda} - \frac{\|W_0\|_1}{\lambda},
\end{align}
by using the disjoint property of $\Omega_E$ and $\Omega_E^c$, we have
\begin{align}
&\|E_0-A\|_1 - \|E\|_1 \notag \\
&= \|E - P_{\Omega_{E}}A - P_{\Omega_{E}}^cA\|_1 - \|E\|_1 \notag \\
&= \|E - P_{\Omega_{E}}A\|_1 + \|P_{\Omega_{E}}^cA\|_1 - \|E\|_1 \notag \\
&\geq \|E\|_1 - \|P_{\Omega_{E}}A\|_1 + \|P_{\Omega_{E}}^cA\|_1 - \|E\|_1 \notag \\
&= \|P_{\Omega_{E}}^cA\|_1 - \|P_{\Omega_{E}}A\|_1 \geq \sigma, 
\end{align}
then it follows that
\begin{align}
\label{last2step}
f(L',E') - f(L,E) = \sigma + \frac{\|W'\|_1 - \|W_0\|_1}{\lambda}.
\end{align}
In addition, since $\lambda \geq \frac{\|W_0\|_1}{\sigma}$, we have
\begin{align}
\label{laststep}
\frac{\|W'\|_1 - \|W_0\|_1}{\lambda} > -\frac{\|W_0\|_1}{\lambda} \geq -\sigma
\end{align}
Plugging (\ref{laststep}) into (\ref{last2step}) yields 
\begin{align}
f(L',E') - f(L,E) > 0,
\end{align}
and therefore Theorem \ref{thm1} is proved. 

\bibliographystyle{plain}
\bibliography{arXiv_version_RoSuRe.bib}

\end{document}